\documentclass{article}

\usepackage[nonatbib,final]{nips_2016}

\usepackage[utf8]{inputenc} 
\usepackage[T1]{fontenc}    
\usepackage{hyperref}       
\usepackage{url}            
\usepackage{booktabs}       
\usepackage{amsfonts}       
\usepackage{nicefrac}       
\usepackage{microtype}      
\usepackage{dsttr}
\usepackage{rtrees}
\usepackage{pstricks}
\usepackage{latexsym}
\usepackage{amsfonts}
\usepackage{amssymb}
\usepackage{graphicx}
\usepackage{multirow}

\title{Bootstrapping incremental dialogue systems:
using   linguistic knowledge  to learn from minimal data}

\author{
  Dimitris Kalatzis, Arash Eshghi, and Oliver Lemon
\thanks{This research is funded by the EPSRC, under  grant number EP/M01553X/1 (BABBLE project: \url{http://sites.google.com/site/hwinteractionlab/babble}). Eshghi and Lemon developed the overall method and wrote the paper. Eshghi is the main developer of DS-TTR and its integration with RL. Kalatzis integrated with  Deep RL and ran the experiments.}\\
  Interaction Lab, Department of Computer Science\\
  Heriot-Watt University,  Edinburgh  \\
  \texttt{dimitriskalatzis89@gmail.com,eshghi.a@gmail.com,o.lemon@hw.ac.uk} \\
}

\begin{document}

\maketitle

\begin{abstract}
  We present a method for inducing    new   dialogue systems from very small amounts of 
unannotated  dialogue data, showing how word-level exploration using Reinforcement Learning (RL), combined with an incremental and semantic grammar - Dynamic Syntax (DS) - allows systems to discover, generate, and understand many new dialogue variants. 
The method avoids the use of expensive and time-consuming dialogue act annotations, and supports more natural (incremental) dialogues than turn-based systems. 
 Here, language generation and dialogue management are treated as a joint decision/optimisation problem, and  the  MDP model for RL is constructed automatically. With an implemented system, we show that
  this method   enables a wide range of dialogue  variations to be automatically captured, even when the system is trained from only a single  dialogue. The variants include question-answer pairs, over- and under-answering, self- and other-corrections, clarification interaction, split-utterances, and ellipsis.
 This generalisation property results   from the  structural knowledge and constraints present within the DS grammar, and highlights some   limitations of recent  systems   built using machine learning techniques only.
\end{abstract}


\section{Introduction} \label{sec:related_work}



Recent data-driven machine learning approaches treat dialogue as a sequence-to-sequence generation problem, and train their models from large datasets, e.g.\  \cite{Wen.etal16a,Wen.etal16b,vinyals}. As a result, while these systems   reproduce patterns found in   training data, they do not exploit any structural knowledge about language encoded in grammars and formal models of dialogue. 
However, the interpretation of many context-dependent utterances in dialogue depends on the underlying structure and content of prior dialogue turns (see e.g.\ 
 \cite{Purver.Ginzburg04,Eshghi.etal15} for how clarification requests are interpreted).
In consequence,   models that rely on surface features of the dialogue alone (i.e.\ words)  
may
have  limitations in handling such data (e.g.\ by providing a relevant response), even if they have observed the relevant sequences   often.
Furthermore, as these systems do not parse to logical forms (i.e.\ a compositional, interpretable representation), they do not allow for inference, and this further limits their application since such a system has no notion of why or how it acts the way it does, and so cannot explain its actions or reasoning.

For these reasons, we explore how formal grammars and dialogue models can be combined with machine learning methods, where  linguistic knowledge is used to   bootstrap new dialogue systems from very small amounts of unannotated data. This also has the important benefit of reducing developer effort. In addition, we learn dialogue policies at the word-level, rather than turn level -- producing  more natural dialogues that are known to be preferred by users (e.g.\ \cite{aistincremental}, and see examples in figure \ref{variation}).

\section{Inducing Dialogue Systems}\label{model}

Our overall method combines incremental dialogue parsing and Reinforcement Learning  for system utterance generation in context.
We employ a Dynamic Syntax (DS) parser \cite{Kempson.etal01}
for incremental language understanding and tracking of the dialogue state using Eshghi et al.'s model of feedback in dialogue \cite{Eshghi.etal15,Eshghi15}, and a set of transcribed successful dialogues $D$ in the target application domain.

To \emph{automatically} construct a Markov Decision Process for $D$ we induce it using DS as described in section \ref{inducing}. We define the state encoding function $F: C \rightarrow S$, where any $c \in C$ is a DS context and $s \in S$ is a (binary) state vector. For more details see section \ref{encoding}. Finally we define the action set as the DS lexicon $L$ (i.e.\ MDP actions are words) and the reward function $R$, which is described in section \ref{rlmethod}.
We then use Reinforcement Learning to train a policy $\pi : S \rightarrow L$, where $L$ is the DS Lexicon, and $s = F(c) \in S$, where $c$ is the (incrementally constructed) dialogue context as output by DS at any point in a  dialogue. The system is trained in interaction with a (semantic) simulated user, also automatically built from the dialogue data -- see section \ref{usersim}.

\noindent
The resulting learned policy forms the combined (incremental) DM and NLG components of a dialogue system for
$D$: i.e.\ a jointly optimised action
  selection mechanism for DM and NLG, with DS providing the language understanding
  component.
We now go into the details of the above steps:

\subsection{Inducing the MDP state space}\label{inducing}
We induce an MDP state space from the relevant semantic features in the dialogue data $D$   by tracking all and only those semantic features which are relevant in that domain.
  These constitute the goal contexts reached in the dialogues in $D$, expressed as Record Types (RT) in Type Theory with Records \cite{Cooper05}; where each feature is in the form of an atomic (i.e.\ non-decomposable) RT - usually a predicate, packaged together with its argument fields (see Fig~\ref{fig:encoding} for example RT features).
Importantly for us here, the standard
\textit{subtype} relation $\subtype$ can also be defined for RTs, and is used in the state encoding function (section \ref{encoding}).

To induce the MDP state space, we parse all $d_i \in D$ using DS, generating a set of final success contexts, $\{c_1,\ldots,c_n\}$.
We take the Maximally Specific Common Supertype (MCS -- see \cite{Hough.Purver14}) and abstract out the domain `slot values'. This process has been dubbed `delexicalization' in recent work \cite{Wen.etal16a,Gasic.etal13}, but we note that while this has previously been  done on the dialogue surface level, either by hand or via an external domain ontology, here we do it automatically.
 This results in the goal contexts for $D$, containing the semantic features to be tracked in the MDP state space. Finally, we proceed to decompose these into their constituent, atomic semantic features that will go on to be encoded by the state encoding function. 


For example, the semantic RT features being tracked in Fig.~\ref{fig:encoding} have resulted from automatically decomposing   goal contexts of $D$ in the consumer electronics domain. From left to right, these correspond roughly to the following: {\it ``there is something that's a brand''; ``there is a liking (or wanting, or all equivalents) event in the present tense''; ``there is something made by that brand''; ``the subject of the liking event is the user'';``the object of the liking event is the thing by that brand''.}

\subsection{The state encoding function}\label{encoding}
\begin{figure}[h]
\centering
\includegraphics[scale=0.28]{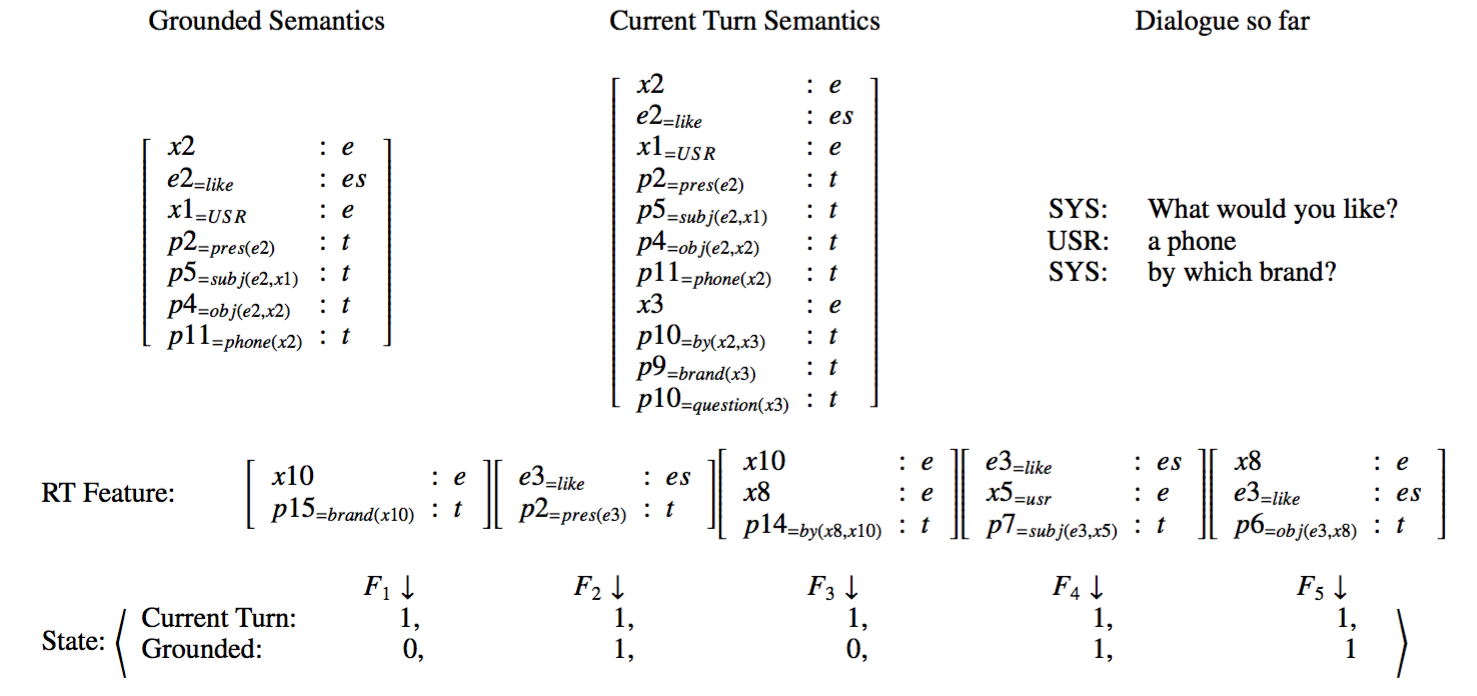}
\caption{Semantics to MDP state encoding with Record Type (RT) features. `Grounded' semantics is the content agreed upon by the participants so far, as computed by DS}\label{fig:encoding}\vspace{-0.1cm}
\end{figure}

As shown in figure \ref{fig:encoding} the MDP state is a binary vector of size $2\times | \Phi |$, i.e.\ twice the number of the RT features. The first half of the state vector contains the grounded features (i.e.\ agreed by the participants)    $\phi_i$, while the second half contains the current semantics being incrementally built in the current dialogue utterance.

Formally the state vector is given by:
$s = \langle F_1(c), \ldots, F_m(c), F_1(c), \ldots, F_m(c)\rangle$; \newline
where $F_i(c)=$ 
    1 if  $c\subtype\phi_i$, and 
    0            otherwise.
(Recall that $\subtype$  is the RT subtype relation).


\subsection{Semantic User Simulation}\label{usersim}
Unlike most other dialogue systems, ours isn't based on dialogue act representations, and is word-by-word incremental. In this setup the notion of a dialogue {\it turn} has no clear definition. Turns are here defined by the user simulator that interrupts system generation according to boundaries encountered in the data.
The rules for interrupting system generation and outputting a user utterance are extracted automatically from the unlabeled dialogue data ($D$) via incremental parsing   using  DS. Intermediate contexts and user utterances occurring in the dialogue are recorded for use by the simulator. As these rules are semantic, they generalise across different interactional variants by assigning user utterances to matched 
 dialogue contexts. If the simulator cannot match the contexts, the system output is considered out-of-domain and is penalised.
Formally, the extracted rules for the simulator are of the form: $sem \rightarrow \{u_1, \ldots, u_n\}$; where $sem$ is the current semantics of some prior system turn, and, the $u_i$ are the utterances (strings) output by the user in that context, as observed in $D$.

\subsection{Reinforcement Learning method}\label{rlmethod}

We have adapted and used Karpathy's \cite{Karpathy15} implementation of Mnih et al.'s \cite{Mnih.etal13} Deep Q-Learning algorithm, to use a deep neural network to estimate the Q-function. 

{\bf The Reward Function} is:
 $R(s) = -1.0$ in the case of out-of-context, ungrammatical, or lengthy utterances; 
$R(s) = 1.0$ when the agent reaches the final (goal) context;  and $R(s) = 0.0$ otherwise.


We have trained policies with this method which are able to successfully perform dialogues in the domain of electronics sales, see figure \ref{variation}.




\section{Evaluation}
We have bootstrapped  a system using this method from only a single dialogue, showing that incremental dialogue systems can be automatically created from small amounts of transcribed dialogue data. 
 Besides reducing development time and cost (since it requires no annotated data), our system discovers (and can process) many interactional variations not found in the training data.
For example, figure \ref{variation}, shows several structural dialogue variants that have been discovered by the system (via RL policy exploration), when it has been given only a single training dialogue. The training dialogue was: {\it ``SYS --  What would you like?; 
    USR --a phone;
    SYS--  by which brand?; 
   USR -- by Apple.''}

  

\begin{figure}[ht]
\begin{footnotesize}
\begin{tabular}{|l|l|l|l|}
\hline
\multirow{5}{7em}{USR: I would like an LG phone\\ SYS: okay.} & 
USR: I would like a phone. & USR: I would like a phone & SYS: what would you like? \\ & 
SYS: by which brand? & SYS: ...by? & USR: a phone\\ &
USR: Apple. & USR: LG. & SYS: ...by?\\ &
USR: okay. & SYS: okay. & USR: Samsung\\ & 
& & SYS: okay.\\
\hline
\multirow{5}{7em}{SYS: you like...?\\ SYS: a phone\\ SYS: ...by?\\ USR: Google\\ SYS: okay.} & 
SYS: what would you like? & SYS: you would like...? & SYS: you like...?\\ & USR: a phone by LG & USR: I would like a computer & USR: a tablet by Google.\\ &
SYS: okay. & SYS: by which brand? & SYS: okay.\\ &
& USR: Apple. & \\ &
& SYS: okay. & \\
\hline
\end{tabular}
\end{footnotesize}
\caption{Some of the interactional variants discovered/learned, from a  single training dialogue.}\label{variation}
\end{figure}

The benefits of such   incremental dialogue variants have been empirically established in prior work (e.g.\ \cite{aistincremental}).
Our work shows the additional benefits of combining linguistic knowledge with machine learning methods: minimising the role of the dialogue engineer, and rapid domain transfer. Ongoing work involves integrating this method with an end-to-end spoken dialogue system framework, and more substantial evaluation with real users. We are also employing this method for the task of learning perceptually grounded language \cite{Yu.etal16}. 

\footnotesize
\bibliographystyle{plain}
\bibliography{all,SG_B,babble}
\end{document}